\documentclass[nohyperref]{article}

\usepackage{microtype}
\usepackage{graphicx}
\usepackage{booktabs} 

\usepackage{hyperref}

\usepackage[accepted]{icml2023}

\usepackage{amsmath}
\usepackage{amssymb}
\usepackage{mathtools}
\usepackage{amsthm}
\usepackage{enumitem}
\usepackage{bbm}
\usepackage{subcaption}

\usepackage[capitalize,noabbrev]{cleveref}

\theoremstyle{plain}

\theoremstyle{definition}

\theoremstyle{remark}

\usepackage[textsize=tiny]{todonotes}
\usepackage{xspace}

\newcommand{\ourmethod}{Sequential Multi-Dimensional Self-Supervised Learning\xspace}
\newcommand{\ourmethodshort}{SMD SSL\xspace}

\icmltitlerunning{\ourmethod for Clinical Time Series}

\begin{document}

\twocolumn[
\icmltitle{\ourmethod for Clinical Time Series}



\icmlsetsymbol{equal}{*}

\begin{icmlauthorlist}
\icmlauthor{Aniruddh Raghu}{MIT}
\icmlauthor{Payal Chandak}{HST}
\icmlauthor{Ridwan Alam}{MIT}
\icmlauthor{John Guttag}{MIT}
\icmlauthor{Collin M. Stultz}{MIT}
\end{icmlauthorlist}
\icmlaffiliation{MIT}{Massachusetts Institute of Technology, Cambridge, MA, USA}
\icmlaffiliation{HST}{Harvard-MIT Program in Health Sciences and Technology, Cambridge, MA, USA}

\icmlcorrespondingauthor{Aniruddh Raghu}{araghu@mit.edu}

\icmlkeywords{Machine Learning, ICML}

\vskip 0.3in
]



\printAffiliationsAndNotice{}  

\begin{abstract}
Self-supervised learning (SSL) for clinical time series data has received significant attention in recent literature, since these data are highly rich and provide important information about a patient's physiological state.
However, most existing SSL methods for clinical time series are limited in that they are designed for unimodal time series, such as a sequence of structured features (e.g., lab values and vitals signs) or an individual high-dimensional physiological signal (e.g., an electrocardiogram).
These existing methods cannot be readily extended to model time series that exhibit multimodality, with structured features \textit{and} high-dimensional data being recorded at each timestep in the sequence.
In this work, we address this gap and propose a new SSL method --- \textit{Sequential Multi-Dimensional SSL} --- where a SSL loss is applied both at the level of the entire sequence and at the level of the individual high-dimensional data points in the sequence in order to better capture information at both scales. Our strategy is agnostic to the specific form of loss function used at each level -- it can be contrastive, as in SimCLR, or non-contrastive, as in VICReg.
We evaluate our method on two real-world clinical datasets, where the time series contains sequences of (1) high-frequency electrocardiograms and (2) structured data from lab values and vitals signs.
Our experimental results indicate that pre-training with our method and then fine-tuning on downstream tasks improves performance over baselines on both datasets, and in several settings, can lead to improvements across different self-supervised loss functions.
\end{abstract}

\section{Introduction}
In clinical settings such as the intensive care unit (ICU), patients are closely monitored and consequently generate a profusion of time series data. This rich data contains significant physiological information about a patient's state and progression over time \citep{johnson2016mimic}. As a result, there have been many efforts to study representation learning and pre-training on these data, particularly using self-supervised learning (SSL) strategies, with the goal of using the pre-trained models for various downstream predictive tasks \citep{mcdermott2021chil, weatherhead22a, tonekaboni2021unsupervised, yeche2021neighborhood,tipirneni2022self}. 

Although these works develop effective strategies to model clinical time series data, they focus only on unimodal time series, such as a sequence of structured features alone, or an individual high dimensional physiological signal.  In reality however, data originating from a patient's encounter is significantly more complex, containing multimodal data recorded at regular intervals. As an example, a given patient may have two very different types of data recorded hourly: (1) high-frequency physiological signals (e.g., an electrocardiogram recorded at 240 Hz); and (2) structured data from labs and vitals signs. These modalities provide complementary information about a patient's physiological state.
Extending existing SSL methods to operate on these time series is challenging, since they do not deal with sequences of high-dimensional data, and do not contend with the multimodal data stream.

In this paper, we take steps towards addressing this gap and outline an approach for self-supervised pre-training on these complex clinical time series. We propose a SSL strategy where we jointly optimize two SSL losses to better capture structure in the data. Our contributions are as follows:
\begin{enumerate}[nosep, leftmargin=*]
    \item We formalize the problem of self-supervised learning (SSL) on \textit{trajectories}, our abstraction of a multimodal time series that contains complex, high-dimensional data recorded at each timestep in the sequence.
    \item We outline a new SSL method, \textit{\ourmethod} (\ourmethodshort), for trajectories. Motivated by the structure of trajectories, \ourmethodshort incorporates two losses: (1) a component SSL loss on the level of individual high dimensional data points in the sequence; and (2) a global SSL loss on the level of the overall sequence. \ourmethodshort can be instantiated with contrastive losses, as in SimCLR \citep{chen2020simple} or non-contrastive losses, as in VICReg \citep{bardes2022vicreg}. This is beneficial since different loss functions may be effective in different applications.
    \item We evaluate \ourmethodshort on two real-world clinical datasets where the time series contains sequences of (1) high-frequency electrocardiograms and (2) structured data from labs and vitals signs. On both datasets and on two downstream tasks --- (1) detecting elevated pulmonary pressures and (2) predicting 24 hour mortality --- we find \ourmethodshort improves performance over baselines. In several settings, we observe performance boosts using both SimCLR and VICReg objective functions.
\end{enumerate}

\section{Related work} 
\textbf{Self-supervised learning (SSL).} SSL methods are used to pre-train models and/or learn generalizable representations using unlabeled data. Many existing methods take either a multiview perspective  \citep{chen2020simple,chen2020improved, bardes2022vicreg,he2020momentum,chen2020improved,grill2020bootstrap} or an autoregressive denoising approach \citep{vincent2008extracting, he2022masked}. Here, we focus on multiview approaches since they have been effective in improving predictive performance on clinical tasks \citep{weatherhead22a, tonekaboni2021unsupervised}. 

\textbf{SSL for clinical data.} Existing applications of multiview SSL to medical data have been focused either on physiological signals, such as electrocardiograms (ECGs) \citep{cheng2020subject,kiyasseh2021clocs,gopal20213kg,diamant2021patient,oh2022lead}, on sequences of tabular data, such as laboratory tests \citep{yeche2021neighborhood,li2021hi}, or on medical imaging \citep{ren2022spatiotemporal}. In contrast, we consider SSL on time series where individual timesteps contain both high-dimensional data (such as ECGs) and structured features. 
Prior studies exploring SSL on multimodal medical data typically infer one modality of data from the other at test time, such as predicting radiologist comments from chest X-ray images \citep{Tiu2022}, rather than modeling sequences of multimodal data where the modalities present non-overlapping sources of information, as we do here.

\textbf{Multilevel SSL loss functions.} Our method uses a two-level loss function that is motivated by the complex structure of the data stream we consider: sequences in which individual elements are themselves high-dimensional. 
A related approach in computer vision, VICRegL \citep{bardes2022vicregl}, applies multilevel self-supervision to images where patch-level similarity is defined using spatial transformations. In contrast to VICRegL, which formulates a component level loss using patches, our method formulates a component level loss on entire signals, and so operates on a different level of abstraction. Another recent method decouples local and global representation learning for a single time series \citep{pmlr-v151-tonekaboni22a}. This work also operates on a different level of abstraction, since it does not consider sequences of time series. Finally, \citet{ren2022spatiotemporal} demonstrate that multiscale SSL for neuroimaging offers improvement on downstream tasks. However, their techniques are tailored for neuroimaging and do not generalize readily to the data we consider.

\begin{figure*}[t]
\centering
\includegraphics[width=\linewidth]{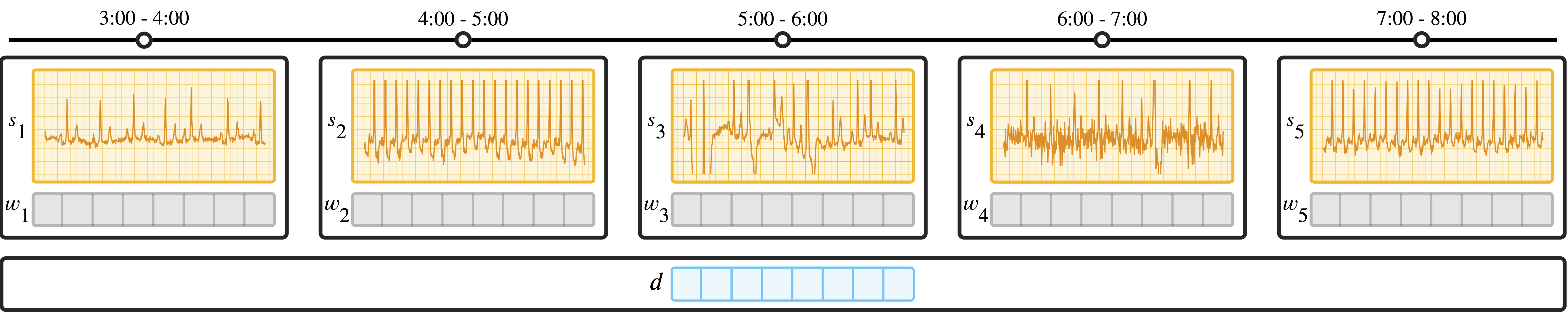}
\caption { \small \textbf{An example of a multimodal clinical time-series or `trajectory'.} The trajectory $\tau$ contains an static vector $d$ consisting of measurements that remain constant over the time period, and a time series of high-dimensional physiological signals $s_t$ and structured data $w_t$ measured at each time step. Here, each time step is a 1 hour window.}
\label{fig:traj}
\end{figure*}

\section{Methods}
\label{sec:methods}
In this section, we describe our approach for self-supervised learning (SSL) on multimodal clinical time series: \textit{Sequential Multi-Dimensional SSL.}
We first outline our problem setup, describing the multimodal data stream that we consider. We then detail our SSL scheme, specifying the loss functions used to learn representations on both an individual timestep (component) level and on a overall sequence (global) level. We conclude with a discussion of other applications of the method. 

\subsection{Problem Setup}
\label{sec:methods:probsetup}
\paragraph{Defining \textit{trajectories}.} We use the term \textit{trajectory} to refer to a sequence of physiological signals and structured data collected over time for a patient. This definition is motivated by an important use-case in cardiovascular medicine, where patients may be monitored with telemetry devices that regularly record physiological waveforms in addition to having lab tests and vitals signs periodically measured. The concept of a trajectory could be readily expanded to include other information, such as imaging or medications, depending on the context.

Formally, a trajectory 
$\tau$ 
of length $T$
has the structure: 
$$\tau = (d, \{(w_t,s_t)\}_{t=1}^{T}).$$ 
Here, $d \in \mathbb{R}^L$ represents a set of static features that do not change over the trajectory (such as demographic information or infrequently measured lab values). The sequence $\{(w_t,s_t)\}_{t=1}^{T}$ contains a vector of structured data $w \in \mathbb{R}^M$, and a high-dimensional signal $s \in \mathbb{R}^{C \times P}$, where $C$ is the number of signal channels and $P$ is the number of samples in the signal (typically on the order of a few thousand).  
A visualization of a trajectory is shown in Figure~\ref{fig:traj}. 

\textbf{Trajectory neural network.} Trajectories are mapped into vector representations using a neural network $f_\theta$ with three components (Figure \ref{fig:arch}): (1) a static and structured features encoder~$f_\theta^{w,d}$; (2) a signals encoder $f_\theta^{s}$; (3) a sequence module~$f_\theta^{\tau}$. At each timestep $t$, the modalities are embedded and concatenated into timestep representations: \mbox{$z_t = \text{concat}\left(f_\theta^{w,d}(w_t,d), f_\theta^{s}(s_t)\right)$}. The sequence module maps these representations into a vector: \mbox{$z = f_\theta^{\tau} (z_1, z_2, \ldots, z_T)$}. In a supervised setting, a classifier $c_\psi$ maps $z$ to a predicted label $\hat y$. 

\textbf{Trajectory self-supervised learning (SSL).} Inspired by recent work in SSL \citep{chen2020simple,chen2020big,bardes2022vicreg,zhang2022self}, we consider a two-stage learning problem: pre-training (PT) followed by fine-tuning (FT). We first pre-train the model $f_\theta$ on an unlabelled dataset of trajectories using some SSL algorithm, and then evaluate the SSL method by FT this pre-trained model on a set of downstream tasks and measuring performance on these tasks. At FT time, different paradigms could be used -- we could initialize a classification head and then fine-tune the whole model, or train a linear classifier on the frozen model.

To pre-train the model, we assume access to an unlabelled PT dataset of $N_{\text{PT}}$ patient trajectories, $\mathcal{D}_{\text{PT}} = \{\tau^{(n)}\}_{n=1}^{N_{\text{PT}}}$. To fine-tune the model, we assume access to a set of labelled FT datasets -- given $K$ FT tasks indexed by $k$, we denote each FT dataset as $\mathcal{D}_{\text{FT}}^{(k)}= \{(\tau^{(n)}, y^{(n)}\}_{n=1}^{N_{\text{FT}}^{(k)}}$, where $y^{(n)}$ denotes the label for a trajectory $\tau^{(n)}$. 

\subsection{Sequential Multi-Dimensional SSL}
\label{sec:method-detail}
\begin{figure*}[t]
\centering
\includegraphics[width=\linewidth]{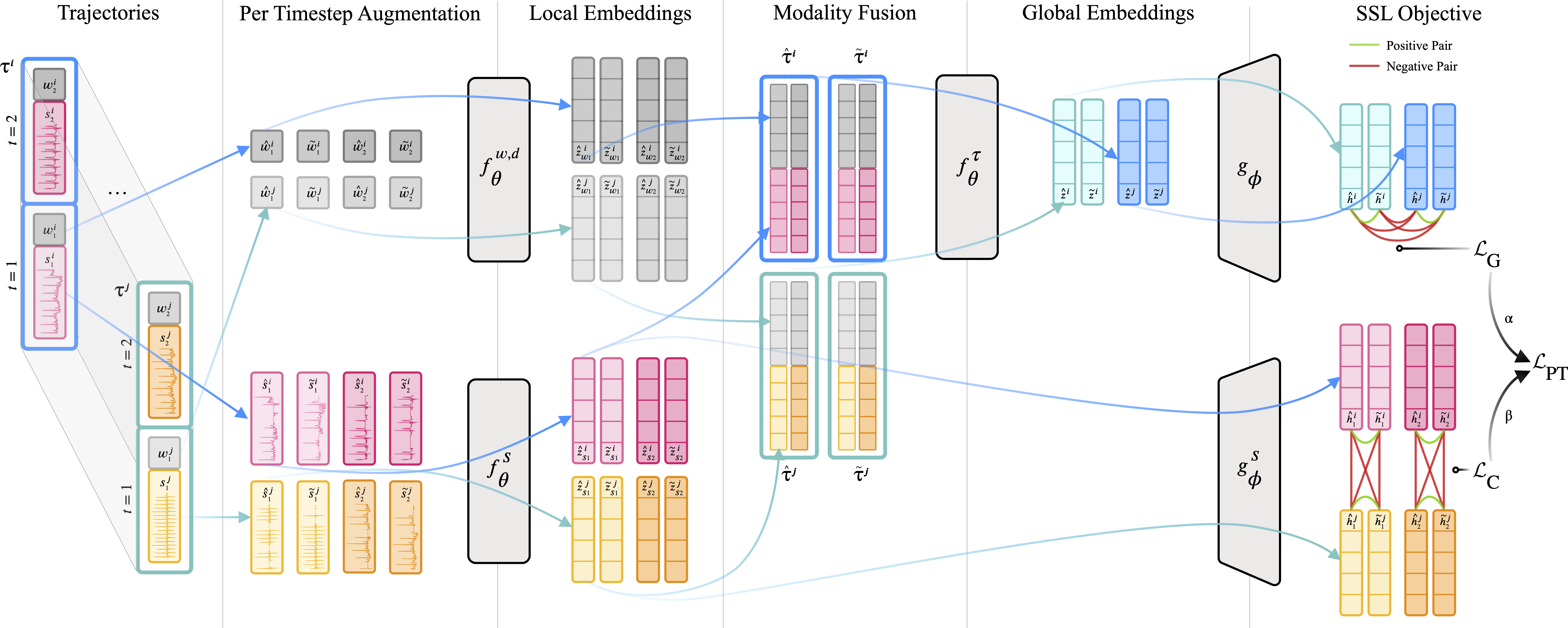}
\caption { \small \textbf{Overview of \ourmethod (\ourmethodshort),} which uses losses at two levels to encourage effective pre-training on complex time series. We start with a batch of trajectories, each denoted $\tau$, consisting of a static vector $d$ (not shown for clarity) and a sequence of signals $s_t$ and structured data $w_t$ (sequence of length 2 here). These data are augmented on a per-modality and per-timestep basis (arrows show flow for the data at a single timestep) and passed through encoders $f_\theta^{s}$ and $f_\theta^{w,d}$ to generate local embeddings of the signals and structured data at each timestep. The signal embeddings pass through a projection head $g_\phi^s$, after which we compute a component SSL loss $\mathcal{L}_{\text{C}}$. Separately, the embedding of the entire trajectory (obtained by concatenating the per-modality embeddings) is passed through a sequence model $f_\theta^{\tau}$ and a global projection head $g_\phi$, on which we compute the global SSL loss $\mathcal{L}_{\text{G}}$. The total loss $\mathcal{L}_{\text{PT}}$ is a weighted sum of the component and global losses. \ourmethodshort can be instantiated with both contrastive and non-contrastive losses -- shown here is a contrastive framing (as in SimCLR) with explicit negative pairs.}
\label{fig:method-overview}
\end{figure*}

We propose a new method for SSL on trajectories -- Sequential Multi-Dimensional SSL (\ourmethodshort), depicted in Figure~\ref{fig:method-overview}.  
Our approach builds on multi-view SSL like SimCLR \citep{chen2020simple} and VICReg \cite{bardes2022vicreg}, since prior work has successfully used these strategies on clinical data  \citep{diamant2021patient,kiyasseh2021clocs,gopal20213kg, vumedaug2021,oh2022lead}. 

\ourmethodshort uses a loss function with two terms --  a global loss, computed at the trajectory level, and a component loss, computed at the individual signal level. We now describe these two losses, and then present the overall objective.

\subsubsection{Global Loss}
The global loss $\mathcal{L}_\text{G}$ encourages the encoding model $f_\theta$ to embed similar trajectories to similar points in the representation space. We follow related work \citep{chen2020simple} and define a similar (or positive) pair of trajectories to be those that are augmentations of the same base trajectory. 

Given a trajectory-level augmentation function, the computation of the global loss proceeds as follows:
\begin{enumerate}[nosep, leftmargin=0.5cm]
    \item Sample a batch of trajectories from the PT dataset: $\{\tau^{(n)}\}_{n=1}^{B}$, where $B$ is the batch size.
    \item For each trajectory $\tau^{(n)}$, generate two augmented views of it: $\tilde{\tau}^{(n)}$ and $\hat{\tau}^{(n)}$.
    \item Pass the augmented views through the representation model $f_\theta$ and a projection head $g_\phi$ generating two sets of projections: $\tilde{h}^{(n)}$ and $\hat{h}^{(n)}$. 
    \item Assemble projected pairs into two sets of matrices: $\hat{H} = [\hat{h}^{(1)}, \ldots, \hat{h}^{(B)}]$, $\tilde{H} = [\tilde{h}^{(1)}, \ldots, \tilde{h}^{(B)}]$. 
    \item The global loss is equal to the trajectory self-supervised loss $\mathcal{L}_{\text{SSL}}^\tau$ computed on these projections:
    \begin{align}
        \mathcal{L}_\text{G} = \mathcal{L}_{\text{SSL}}^\tau(\hat{H}, \tilde{H}) \label{eqn:global}.
    \end{align}
\end{enumerate}
The choice of the trajectory self-supervised loss $\mathcal{L}_{\text{SSL}}^\tau$ is a design decision; one choice is the normalized temperature-scaled cross-entropy loss (NT-Xent) as in SimCLR \citep{chen2020simple, chen2020big}. Given all $2B$ positive pairs $(h_i, h_j)$ as the rows of the two matrices $[\hat{H},\tilde{H}]$ and $[\tilde{H}, \hat{H}]$, we compute:
\begin{align}
    \mathcal{L}_{\text{NT-Xent}}^\tau &= \frac{1}{2B}\sum_{i=1}^{2B} -\log\frac{\exp(\mathrm{sim}(h_i, h_j)/\gamma)}{\sum_{k=1}^{2B} \mathbbm{1}_{[k \ne i]}\exp(\mathrm{sim}(h_i, h_k)/\gamma)} \label{eqn:simclr},
\end{align} 
    where $\mathrm{sim}(a, b)$ 
    is cosine similarity and $\gamma$ is the temperature hyperparameter.
Another choice is the VICReg loss, minimizing mean squared error between positive pairs, with variance and covariance regularizers \citep{bardes2022vicreg}:
\begin{align}
    \mathcal{L}_{\text{VICReg}}^\tau = \lambda I(\hat{H}, \tilde{H}) + \mu {\text{Var}}(\hat{H}, \tilde{H}) + \nu {\text{Cov}}(\hat{H}, \tilde{H}) \label{eqn:vicreg},
\end{align}
    where $I()$ is the mean squared error, $\text{Var}()$ is the variance regularizer, $\text{Cov}()$ is the covariance regularizer, and $\lambda, \mu,$ and $\nu$ are hyperparameters.
Flexibility in the form of the loss function is beneficial since different applications might benefit from different losses. In our experiments, we focus on the NT-Xent and VICReg loss functions. 

\subsubsection{Component Loss} 
Pre-training with the global loss is a straightforward application of SSL to trajectories. However, each trajectory contains complex substructures (the high-frequency signals $s_t$) and the global loss alone may not be sufficient to guide the model to learn useful representations of these substructures. 
We hypothesize that incorporating a second loss term on the individual signal level, the \textit{component loss} $\mathcal{L}_\text{C}$, would lead to learning richer representations of the signals. 

Given a signal augmentation function, we compute the component loss as follows: 
\begin{enumerate}[nosep, leftmargin=0.5cm]
    \item Sample a batch of trajectories from the PT dataset: $\{\tau^{(n)}\}_{n=1}^{B}$, where $B$ is the batch size. 
    \item Generate two augmented views of each signal in each trajectory. For a given trajectory $\tau^{(n)}$, let the two augmented sets of signals be: $\{ \tilde{s}_t^{(n)} \}_{t=1}^{T}$ and $\{ \hat{s}_t^{(n)} \}_{t=1}^{T}$.
    \item Pass the augmented views through the signal encoder model $f_\theta^{s}$ and a signal projection head $g_\phi^{s}$ generating two sets of projections: $\{ \tilde{h}_t^{(n)} \}_{t=1}^{T}$ and $\{ \hat{h}_t^{(n)} \}_{t=1}^{T}$. 
    \item Assemble these pairs of projections into pairs on a per-timestep basis: $\hat{S}_t = [\hat{h}_t^{(1)}, \ldots, \hat{h}_t^{(B)}]$, $\tilde{S}_t = [\tilde{h}_t^{(1)}, \ldots, \tilde{h}_t^{(B)}], \forall t= 1, \ldots, T$.
    \item The component loss is equal to the signal self-supervised loss $\mathcal{L}_{\text{SSL}}^s$ averaged over timesteps: 
    \begin{align}
        \mathcal{L}_\text{C} = \frac{1}{T} \sum_{t=1}^T \mathcal{L}_{\text{SSL}}^s (\hat{S}_t, \tilde{S}_t) \label{eqn:local}.
    \end{align}
\end{enumerate}

As with the global loss, the form of the signal SSL loss used is a design decision. 
The intuition for computing the signal SSL loss separately at each timestep is that nearby timesteps in a trajectory can be very similar. Particularly if we use a contrastive loss such as NT-Xent, we do not want these nearby timesteps to serve as negative examples in the contrastive loss. Separating out the computation over timesteps addresses this issue. 

\subsubsection{Overall Objective}
The overall objective used at PT is:
\begin{align}
    \mathcal{L}_{\text{PT}} = \alpha \mathcal{L}_{\text{G}} + \beta \mathcal{L}_{\text{C}} \label{eqn:multilevel}.
\end{align}
The hyperparameters $\alpha$ and $\beta$ control the contributions of the global and component losses. 
Fixing $\alpha = 0$ is SSL on a signal-level alone (only PT the signal encoder) and fixing $\beta=0$ is SSL on the overall trajectory level alone; we evaluate both in our experiments, finding that combining the two losses is beneficial to performance. In Appendix \ref{app:sec:results}, we study the evolution of the two losses over \ourmethodshort training, which provides intuition as to the effect of each term during pre-training.


\subsection{Augmentation Functions}
\ourmethodshort requires augmentations for trajectories and signals in order to compute the global and component losses respectively. We now describe these.

\textbf{Trajectory augmentation.} We form an augmented trajectory by separately augmenting each of the data modalities within the trajectory, using the following approach for each data type (further details in Appendix \ref{app:sec:aug}):
\begin{itemize}[nosep, leftmargin=*]
    \item \textbf{High-frequency signal $s$:} For each signal in the trajectory of length $T$, we form a pair of augmented views by first splitting the signal into two disjoint segments (e.g., taking the first 10 seconds as one view, and the second 10 seconds as the second view) and then applying random masking and noise addition as augmentations to each view independently, similar to the approach used in CLOCS \citep{kiyasseh2021clocs}. The intuition is that two segments of a signal that are close in time should encode similar physiology, and can therefore be considered paired views. Random masking and noise addition are commonly used as time-series augmentations \citep{gopal20213kg, zhang2022self,raghu2022data, iwana2021empirical}.
    \item \textbf{Structured-time series data $w$:} The tabular data sequence forms a $T \times M$ matrix over all timesteps of the trajectory. Following prior work \citep{yeche2021neighborhood}, we apply two data augmentation strategies to this matrix: Gaussian noise addition and history cutout.
    \item \textbf{Static features $d$:} Following \citet{yeche2021neighborhood}, we use random dropout and noise addition. Other corruption strategies (e.g., \citet{bahri2021scarf}), were found to be less effective, potentially due being too strong  (also seen in \citet{levin2022transfer}).
\end{itemize}
We note that our approach of forming augmented trajectories by independently transforming each individual data type is straightforward, but not necessarily optimal. Exploring other strategies for generating multiple views of trajectory data is an important direction of future work.

\textbf{Signal augmentation.} For computational efficiency, we re-use the augmentated signals already generated during the trajectory-level augmentation when computing the component loss.  However, additional/different augmentations could be applied on the signal level.

\subsection{Broader Applications of \ourmethodshort}
We have instantiated \ourmethodshort for a setting in which multimodal trajectories consists of a sequence of structured data and high-dimensional signals. 
More generally, our approach is valuable in any setting where we have sequences of high-dimensional data -- the two-level loss function encourages representation learning on both an individual signal level and an overall sequence level. 
For example, \ourmethodshort could be useful in modeling sequence of medical images for a patient taken over time. The component loss encourages learning rich embeddings of individual images, and the global loss encourages learning temporal trends.

\section{Experiments}
In this section, we evaluate \ourmethod (\ourmethodshort) on two clinical datasets \footnote{Code at \url{https://github.com/aniruddhraghu/smd-ssl}.}.
We begin by describing the datasets, tasks, and experimental setup. We then evaluate \ourmethodshort and baselines in two settings: \textit{unimodal}, with trajectories that contain only a sequence of physiological signals; and \textit{multimodal}, with trajectories containing both signals and structured data. We find that \ourmethodshort performs strongly in both settings. We also analyze \ourmethodshort's sensitivity to the component loss weight and its learned representations. 

\begin{table*}[h]
\centering
\caption{\small Dataset Statistics.}
\begin{tabular}{@{}lcccccc@{}}
\toprule
                                    & \multicolumn{3}{c}{Dataset 1}                                   & \multicolumn{3}{c}{Dataset 2}              \\ \midrule
\multicolumn{1}{l|}{Task}           & \# Patients & \# Trajectories & \multicolumn{1}{c|}{Prevalence} & \# Patients & \# Trajectories & Prevalence \\ \midrule
\multicolumn{1}{l|}{Pre-training}   & 8888        & 43858           & \multicolumn{1}{c|}{N/A}        & 5022        & 26615           & N/A        \\
\multicolumn{1}{l|}{Elevated mPAP}  & 2025        & 48511           & \multicolumn{1}{c|}{77.5\%}     & 500         & 14957           & 87.9\%     \\
\multicolumn{1}{l|}{24hr Mortality} & 9605        & 57758           & \multicolumn{1}{c|}{1.4\%}      & 5689        & 318306          & 2.3\%      \\ \bottomrule
\end{tabular}
\label{tab:dataset}
\end{table*}

\subsection{Datasets and Tasks} 
\label{sec:datasets-tasks}
We consider two clinical datasets (Table \ref{tab:dataset}):
\begin{itemize}[nosep,leftmargin=*]
    \item \textbf{Dataset 1} is a private dataset from Massachusetts General Hospital (MGH), consisting of 9605 patients with a prior diagnosis of heart failure. 
    \item \textbf{Dataset 2} is a public dataset derived from the commonly used MIMIC-III clinical database \citep{johnson2016mimic, Goldberger2000PhysioBankPA} and its associated database of physiological signals \citep{Moody2020-iq}. We use the preprocessing pipeline introduced in \citet{harutyunyan2019multitask} to form our cohort of 5689 patients and extract the structured data features.
\end{itemize}

\textbf{Constructing PT and FT sets.} Each dataset consists of a number of patient hospital visits. We resample each patient's hospital visit at hourly resolution, and each hour of a patient's stay represents a single timestep in our trajectory abstraction. For simplicity, we fix the length of trajectories to be 8 elements (letting a trajectory correspond to a common shift length of 8 hours). 

To generate PT trajectories from these resampled visits, we first split each visit into non-overlapping contiguous 12 hour blocks. A PT trajectory is formed by first sampling a 12 hour block, and then selecting 8 contiguous timesteps from that block (with the starting timestep selected randomly).  We discuss the implications of this (particularly relating to negative samples in contrastive losses) in Appendix \ref{app:sec:dataset_details}.

To generate FT trajectories, we use a sliding window to select contiguous 8 hour blocks at 1 hour increments from each visit. Each of these contiguous 8 hour blocks is a trajectory in the FT dataset. The trajectory labels are formed based on the specific task, as described below.

\textbf{Fine-tuning tasks.} We consider two predictive tasks:
\begin{itemize}[nosep, leftmargin=*]
    \item \textbf{Elevated mPAP}: Each hour, detect whether a patient's mean Pulmonary Arterial Pressure (mPAP) is abnormally high. This task is of clinical interest since the mPAP is typically measured via an invasive study, and so inferring whether it is abnormal using minimally invasive signals (i.e., the ECG, labs, and vitals signs) is valuable. 
    Prior work \citep{rhcnet,raghu2023ecg} studied similar tasks of predicting hemodynamic variables from the 12-lead ECG, but not in the context of online trajectory data, as we do here. 
    \item \textbf{24hr Mortality:} Each hour, predict whether the patient is going to die in the next 24 hours. This task is commonly used to evaluate predictive models for ICU time series data \citep{mcdermott2021chil,yeche2021neighborhood} -- our goal with studying this task is to understand how our approach performs when compared to other established methods. In Dataset 2, this task is named `Decompensation' in the preprocessing pipeline from \citet{harutyunyan2019multitask}; we refer to it at 24hr mortality here. 
\end{itemize}

\textbf{Trajectory features and preprocessing.} 
The trajectories in PT and FT sets consist of static features $d$ and a time-series of structured data and physiological signals $\{(w_t,s_t)\}_{t=1}^{T}$, as presented in Section \ref{sec:methods:probsetup}. The static features $d$ contain information on infrequently measured vitals signs and lab values; $d \in \mathbb{R}^9$ in Dataset 1 and $d \in \mathbb{R}^{38}$ in Dataset 2. At each timestep, $w_t$ captures summary statistics related to regularly measured vitals signs within a 1 hour window; $w_t \in \mathbb{R}^{30}$ in Dataset 1 and $w_t \in \mathbb{R}^{13}$ in Dataset 2. $s_t$ is a 10 second electrocardiogram (ECG) signal extracted from a longer signal measured within each 1 hour window; in Dataset 1, $s_t \in \mathbb{R}^{4\times2400}$ is a 240 Hz 4-channel ECG, and in Dataset 2, $s_t \in \mathbb{R}^{1\times1250}$ is a 125 Hz 1-channel ECG.

Missing structured data are forward-fill imputed if part of a time series and otherwise imputed with the mean over the training dataset. Missing signals are represented with zeros. Any trajectories that have more than 1 timestep with a missing signal are excluded. Further dataset and preprocessing details are in the appendix.

\begin{table*}[t]
\centering
    \caption{\small \textbf{Pre-training with the \ourmethodshort objective improves performance on both datasets in the unimodal setting.} Mean and 95\% confidence interval of AUROC on FT tasks.  We observe that PT methods outperform not doing PT, and training with the \ourmethodshort (component and global) losses boosts performance the most.}
     \label{tab:results-unimodal}
    \begin{subtable}[h]{0.5\textwidth}
    \caption{\small Results on Dataset 1.}
        \label{tab:dataset1-unimodal}
    \centering
        \begin{tabular}{@{}lcc@{}}
            \toprule
                         & Elevated mPAP & 24hr Mortality\\ \midrule
            RandInit           & 65.0 $\pm$ 0.1   & 56.1 $\pm$ 0.6   \\
            SimCLR (global)      & 68.1 $\pm$ 0.1   & 66.7 $\pm$ 0.5   \\
            VICReg (global)      & 66.6 $\pm$ 0.1   & 64.9 $\pm$ 0.5\\
            SimSiam (global)      & 63.8 $\pm$ 0.1   & 50.6 $\pm$ 0.6\\
            SimCLR (component)      &  67.5 $\pm$ 0.1   &  71.7 $\pm$ 0.5   \\
            VICReg (component)      &  68.7 $\pm$ 0.1    & 63.5 $\pm$ 0.4  \\ 
            SimSiam (component)      & 64.2 $\pm$ 0.1   &  54.3 $\pm$ 0.5 \\ \midrule
            \ourmethodshort (SimCLR) & \textbf{69.9 $\pm$ 0.1}   & 72.3 $\pm$ 0.4 \\
            \ourmethodshort (VICReg) & 67.6 $\pm$ 0.1   & \textbf{74.6 $\pm$ 0.5}  \\ \bottomrule
        \end{tabular}
    \end{subtable}%
    \begin{subtable}[h]{0.5\textwidth}
    \caption{\small Results on Dataset 2.}
        \label{tab:dataset2-unimodal}
    \centering
        \begin{tabular}{@{}lcc@{}}
        \toprule
                     & Elevated mPAP & 24hr Mortality \\ \midrule
        RandInit     &  63.4 $\pm$ 0.4  &      54.6 $\pm$ 0.2         \\
        SimCLR (global)       &  65.9 $\pm$ 0.4             &    56.6 $\pm$ 0.2           \\
        VICReg (global)       &  66.7 $\pm$ 0.4             &  53.6 $\pm$ 0.2             \\
        SimSiam (global)      & 61.9 $\pm$ 0.4              & 61.1 $\pm$ 0.2\\
        SimCLR (component)       &   65.7 $\pm$ 0.4            &   61.1 $\pm$ 0.2           \\
        VICReg (component)       &   66.7 $\pm$ 0.4            &   63.1 $\pm$ 0.2\\
        SimSiam (component)      & 65.9 $\pm$ 0.4   &  50.3 $\pm$ 0.2 \\ \midrule
        \ourmethodshort (SimCLR) &  \textbf{67.0 $\pm$ 0.4}     &   \textbf{65.9 $\pm$ 0.2} \\
        \ourmethodshort (VICReg) &  66.6 $\pm$ 0.4             &   58.5 $\pm$ 0.2            \\ \bottomrule
        \end{tabular}
     \end{subtable}
\end{table*}

\subsection{Experimental Setup} 
\textbf{Dataset splits.} We split Dataset 1 on a per-patient level into 80/20 development/test sets and use 20\% of the development set as a validation set.  For Dataset 2, we use the predefined development/test split defined in the preprocessing pipeline \citep{harutyunyan2019multitask}, and use 20\% of the development set (splitting on a per-patient basis) as a validation set.

\textbf{Model architecture.} Recall that the encoder $f_\theta$ has three components: we implement the structured features encoder as a 2-layer MLP, the signals encoder as a 1-D ResNet18 CNN \citep{he2016deep}, and the sequence model as a 4-layer GRU.
We use a 2-layer MLP for the projection head $g_\phi$. The model architecture is described more fully in Appendix \ref{app:sec:exptsetup}.

\textbf{Training setup.} We conduct pre-training for 15 epochs, using a batch size of 128, with the Adam optimizer \citep{kingma2014adam}. We found that model performance did not improve with longer pre-training times (Appendix \ref{app:sec:results}).

At fine-tuning time, we consider two evaluation strategies:
\begin{enumerate}[nosep, leftmargin=*]
    \item Linear evaluation: train a linear classifier on the frozen representations from $f_\theta$ \citep{chen2020simple}.
    \item Full FT: initialize a new linear layer after $f_\theta$ and fine-tune the entire model for a maximum of 10 epochs with Adam (performance did not improve after this point), with early stopping based on validation AUROC.
\end{enumerate}
For each method and task, we report the test set AUROC from the evaluation strategy that obtains the best validation set AUROC.  We adopt this approach since our goal is determine to which self-supervised pre-training approach is best -- in order to do so fairly, we compare results under the evaluation strategy that obtains the highest performance.  To obtain error bars, we use bootstrapping: we sample with replacement 100 bootstraps from the testing dataset, and report the 95\% confidence interval in AUROC over these bootstraps.  Since the mortality task has low prevalence, we additionally check trends in AUPRC in Appendix \ref{app:sec:results}.

\textbf{Unimodal and Multimodal evaluation.} \ourmethodshort is generally applicable when we have sequence-structured data where elements of the sequence are themselves high-dimensional. We consider two instantiations of such sequences here: (1) the \textit{unimodal setting}, where the input trajectory only contains the signals sequence of the input, $\tau = \{s_t\}_{t=1}^{T}$; (2) the full \textit{multimodal setting}, where the input trajectory contains the full input sequence of structured data and signals, $\tau = (d, \{(w_t,s_t)\}_{t=1}^{T})$.

\begin{table*}[t]
\centering
\caption{\small \textbf{Pre-training with the \ourmethodshort objective improves performance in three settings in a multimodal evaluation.} Mean and 95\% confidence interval of AUROC on FT tasks. In all settings except 24hr Mortality on Dataset 1, we observe that \ourmethodshort obtains the best performance. The 24 hour mortality task on Dataset 1 appears to benefit little from incorporating the high-dimensional signals, which could explain why \ourmethodshort does not improve performance here.}
     \label{tab:results-multimodal}
    \begin{subtable}[h]{0.5\textwidth}
    \centering
    \caption{\small Results on Dataset 1.}
        \label{tab:dataset1-multimodal}
        \begin{tabular}{@{}lcc@{}}
        \toprule
                     & Elevated mPAP & 24hr Mortality \\ \midrule
        RandInit (signals)&  65.0 $\pm$ 0.1 & 56.1 $\pm$ 0.6 \\
        SSL (signals) & 69.9 $\pm$ 0.1 & 74.6 $\pm$ 0.5 \\ 
        RandInit (structured)&  65.3 $\pm$ 0.1 & \textbf{79.1 $\pm$ 0.4} \\
        SSL (structured) & 66.7 $\pm$ 0.1 & 79.0 $\pm$ 0.4 \\ \midrule
        RandInit          & 69.1 $\pm$ 0.1   & 79.0 $\pm$ 0.4  \\
        SimCLR (global)      & 69.8 $\pm$ 0.1   & 76.4 $\pm$ 0.5   \\
        VICReg (global)      & 69.4 $\pm$ 0.1   & 78.0 $\pm$ 0.4   \\
        SimSiam (global)    &  69.6 $\pm$ 0.1   & 78.4  $\pm$ 0.4    \\
        SimCLR (component)      &  71.4 $\pm$ 0.1  &  78.6 $\pm$ 0.4  \\
        VICReg (component)      &  64.0 $\pm$ 0.1  &  74.0 $\pm$ 0.5  \\ 
        SimSiam (component)      & 68.4 $\pm$ 0.1  &  79.0 $\pm$ 0.4    \\         \midrule
        \ourmethodshort (SimCLR) & \textbf{72.3 $\pm$ 0.1}   & 77.4 $\pm$ 0.4   \\
        \ourmethodshort (VICReg) & 70.3 $\pm$ 0.1   & 77.0 $\pm$ 0.4   \\ \bottomrule
        \end{tabular}
    \end{subtable}%
    \begin{subtable}[h]{0.5\textwidth}
    \centering
    \caption{\small Results on Dataset 2.}
        \label{tab:dataset2-multimodal}
        \begin{tabular}{@{}lcc@{}}
            \toprule
                         & Elevated mPAP & 24hr Mortality \\ \midrule
            RandInit (signals) & 63.4 $\pm$ 0.4 & 54.6 $\pm$ 0.2 \\
            SSL (signals)&  67.0 $\pm$ 0.4 & 65.9 $\pm$ 0.2  \\
            RandInit (structured)&  65.3 $\pm$ 0.3 & 90.0 $\pm$ 0.1 \\
            SSL (structured) & 68.3 $\pm$ 0.3 & 89.3 $\pm$ 0.1 \\ \midrule
            RandInit     &   65.3 $\pm$ 0.3 &  87.8 $\pm$ 0.1             \\
            SimCLR (global)      &   63.7 $\pm$ 0.4 &  86.6 $\pm$ 0.1             \\
            VICReg (global)      &   70.4 $\pm$ 0.4 &  87.8 $\pm$ 0.1  \\
            SimSiam (global)      &  60.6 $\pm$ 0.3   & 90.4 $\pm$ 0.1   \\
            SimCLR (component)      &   59.7 $\pm$ 0.4 &  89.8 $\pm$ 0.1             \\
            VICReg (component)      &   67.1 $\pm$ 0.4 &  84.4 $\pm$ 0.1  \\ 
            SimSiam (component)      &  67.4 $\pm$ 0.4   & 90.6 $\pm$ 0.1   \\       \midrule
            \ourmethodshort (SimCLR) &   69.9 $\pm$ 0.3 &  88.1 $\pm$ 0.1            \\
            \ourmethodshort (VICReg) &   \textbf{71.6 $\pm$ 0.3} &  \textbf{90.7 $\pm$ 0.1}  \\ \bottomrule
            \end{tabular}
     \end{subtable}
\end{table*}

\textbf{Baselines and \ourmethodshort variations.} Existing methods for SSL on clinical data are not exactly applicable, since they do not study pre-training pipelines for multimodal and multi-dimensional time series; e.g., Neighbourhood Contrastive Learning (NCL) \citep{yeche2021neighborhood} is primarily designed for structured data time series alone, and CLOCS \citep{kiyasseh2021clocs} and SACL \citep{cheng2020subject} operate on single physiological waveforms (rather than sequences of waveforms). 

As a result, we focus in the main paper on evaluating general SSL methods as baselines (with further baselines in Appendix \ref{app:sec:results}), varying whether we use the component and/or global loss, in both unimodal and multimodal settings. Our goal is to understand whether the two-level loss formulation boosts performance. The full set of baselines is as follows (further details in Appendix \ref{app:sec:exptsetup}):
\begin{itemize}[nosep,leftmargin=*]
    \item \textbf{RandInit}: A standard baseline: train a model from random initialization on each FT task.
    \item \textbf{SimCLR (global) \citep{chen2020simple}:} Pre-train using the NT-Xent global loss alone.
    This is SimCLR PT on the trajectory level, setting $\alpha=1, \beta=0$ in Eqn. \ref{eqn:multilevel}.
    \item \textbf{VICReg (global) \citep{bardes2022vicreg}:}  Pre-train using the VICReg global loss alone. This is VICReg PT on the trajectory level, setting $\alpha=1, \beta=0$ in Eqn. \ref{eqn:multilevel}.
    \item \textbf{SimSiam (global) \citep{chen2021exploring}:}  Pre-train using SimSiam at a global level. This is SimSiam PT on the trajectory level.
    \item \textbf{SimCLR (component):} Pre-train using the NT-Xent component loss alone ($\alpha=0, \beta=1$ in Eqn. \ref{eqn:multilevel}).
    \item \textbf{VICReg (component):}  Pre-train using the VICReg component loss alone($\alpha=0, \beta=1$ in Eqn. \ref{eqn:multilevel}).
    \item \textbf{SimSiam (component):}  Pre-train using SimSiam at component level. 
\end{itemize}
We consider two variations of \ourmethodshort:
\begin{itemize}[nosep,leftmargin=*]
    \item \textbf{\ourmethodshort (SimCLR):} Pre-train using \ourmethodshort with the NT-Xent loss (Eqns. \ref{eqn:simclr} and~\ref{eqn:multilevel}), fixing the global loss weight $\alpha=1$ and tuning the component loss weight $\beta$. 
    \item \textbf{\ourmethodshort (VICReg):} Pre-train using \ourmethodshort with the VICReg loss (Eqns. \ref{eqn:vicreg} and \ref{eqn:multilevel}), fixing the global loss weight $\alpha=1$ and tuning the component loss weight $\beta$. 
\end{itemize}

\textbf{Hyperparameters.} There are various hyperparameters to tune, such as learning rates and loss weighting for VICReg and \ourmethodshort.
Evaluating many hyperparameters is computationally expensive (involves doing both PT and FT runs), so we conduct a reduced search on a subset of the hyperparameters.
We include full details in Appendix \ref{app:sec:exptsetup}.

\subsection{Results}
\subsubsection{Unimodal Evaluation}
Table \ref{tab:results-unimodal} shows results in the unimodal setting. We highlight three key takeaways from these results:
\begin{enumerate}[nosep, leftmargin=*]
    \item \textbf{Pre-training (PT) helps performance.} In the unimodal setting, PT (particularly SimCLR or VICReg variants in Tables \ref{tab:dataset1-unimodal} and \ref{tab:dataset2-unimodal}) almost always improve performance over not doing any PT (RandInit in Tables \ref{tab:dataset1-unimodal} and \ref{tab:dataset2-unimodal}). This result is expected, since we would expect that given the complex input space, a PT phase for the highly-parameterized CNN encoder and GRU should condition the model better for the downstream tasks, given the limited amount of labelled data on these tasks. 
    \item \textbf{\ourmethodshort obtains the best performance.} On both datasets, we observe that a \ourmethodshort method does best, suggesting the utility of a two-level loss, which encourages the learning of informative representations on both a signal-level and a sequence-level.
    \item \textbf{\ourmethodshort vs single-level SSL.} When using SimCLR, we find that \ourmethodshort consistently improves on a component-only or global-only SimCLR model. With VICReg, improvements are less consistent, and the component-only VICReg variation often performs the best. This may be because VICReg has many loss weighting terms, and these were not jointly tuned with the component loss weight in \ourmethodshort. A more thorough hyperparameter search, perhaps using efficient gradient-based methods \citep{raghu2021meta}, might improve performance of \ourmethodshort (VICReg). 
\end{enumerate}

\subsubsection{Multimodal Evaluation}
Table \ref{tab:results-multimodal} shows results in the multimodal setting, where the input trajectories consist of both signals and structured data. In addition to the aforementioned baselines, we also include results from RandInit and SSL methods trained on only signals and on only structured data.

We highlight some key takeaways from these results:
\begin{enumerate}[nosep, leftmargin=*]
    \item \textbf{The effect of multimodal data is task-specific.} Incorporating both the signals and structured data leads to improvements in the Elevated mPAP task on both datasets (particularly with \ourmethodshort), but has less significant effects in the 24hr Mortality task. This is likely because the structured data in their raw (relatively low-dimensional) form are highly predictive of mortality, and so there is little benefit to be gained from PT. This is seen clearly when comparing the performance of RandInit (structured) and SSL (structured). This phenomenon has been observed in prior work \citep{mcdermott2021chil}.
    \item \textbf{\ourmethodshort performs effectively.} On the Elevated mPAP task (both datasets) and 24hr mortality task (Dataset 2), \ourmethodshort, either with the SimCLR or VICReg loss function, obtains the best performance among all methods.
    \item \textbf{\ourmethodshort vs single-level SSL.} When compared to single-level SSL, we observe improvements when using \ourmethodshort on Elevated mPAP for both SimCLR and VICReg. However, this is not the case for 24hr Mortality -- for example, component-only SSL with SimCLR on this task performs better than \ourmethodshort. This may arise because structured data PT does not improve performance significantly, and so it is preferable to only pre-train the signals encoder rather than the entire model. 
\end{enumerate}

\subsubsection{Further Results}
\textbf{Additional evaluation.} In Appendix \ref{app:sec:results}, we present three additional experiments: (1) comparing \ourmethodshort to other global-only baselines using different loss functions (NCL, CLOCS, and SACL), finding that \ourmethodshort improves on these methods; (2) comparing linear evaluation to full FT for different methods, finding that full FT improves on linear evaluation; and (3) studying the effect of longer PT times, finding that performance does not improve.

\begin{figure}[t]
\centering
\includegraphics[width=\linewidth]{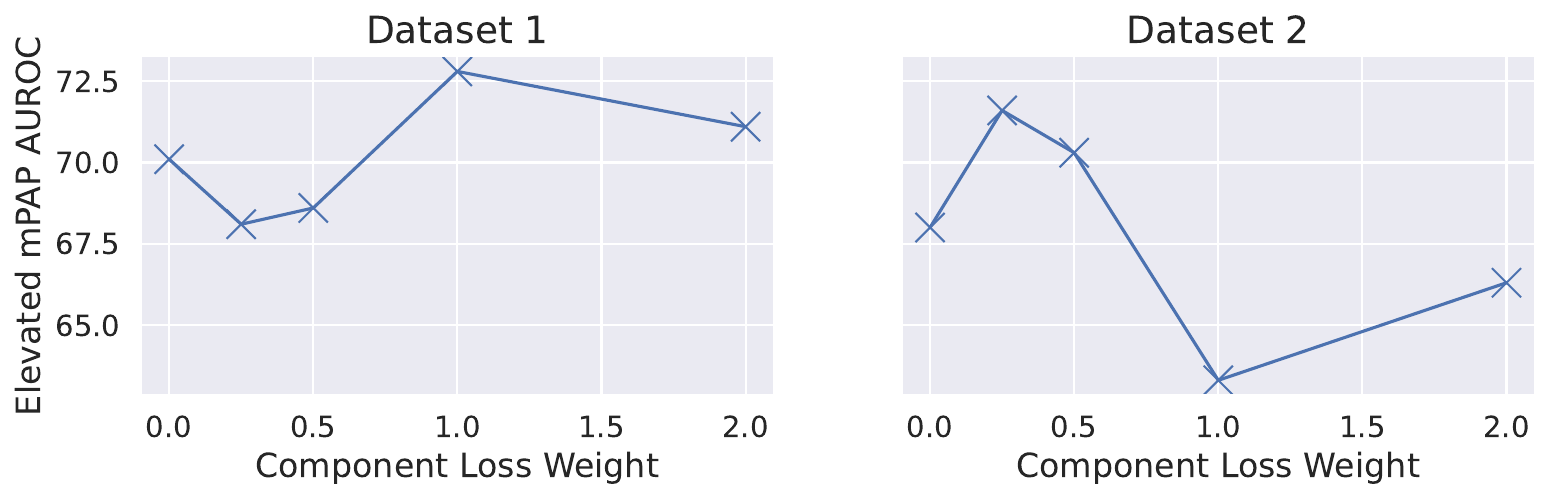}
\caption { \small \textbf{Studying sensitivity to the component loss weight.} We find a higher optimal loss weight on Dataset 1 compared to Dataset 2, possibly due to Dataset 1 having more complex signals.}
\label{fig:loc_loss}
\end{figure}

\textbf{Component loss weight sensitivity.} Considering \ourmethodshort (SimCLR) in the unimodal setting, we fix the global loss weight $\alpha = 1.0$ and vary the component loss weight $\beta$ for the Elevated mPAP task. The validation set AUROC results are shown in Figure \ref{fig:loc_loss}. The optimal value of the component loss weight is higher for Dataset 1, perhaps because the signals in Dataset 1 are more complex than the signals in Dataset 2 (higher sampling rate, multiple channels).

\begin{figure}[t]
\centering
\includegraphics[width=0.75\linewidth]{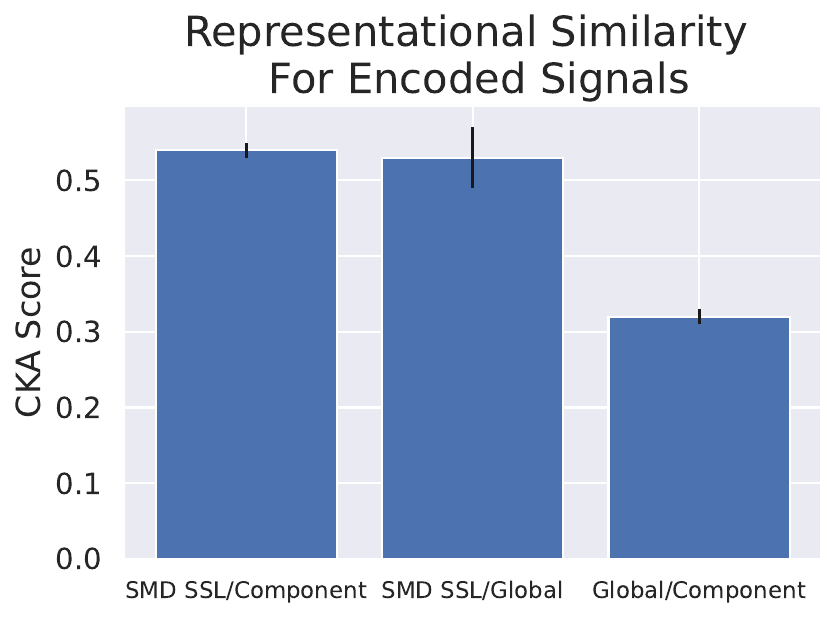}
\caption { \small \textbf{Learned representations from \ourmethodshort are similar to both component-only and global-only SSL.} Comparing learned representations of the signals using different SimCLR-based SSL strategies -- \ourmethodshort, Global, and Component, using Centered Kernel Alignment (CKA). Global and Component SSL show low representational similarity (right-most bar) but \ourmethodshort shows higher similarity with both individually, suggesting that learned representations in \ourmethodshort encode aspects of both component and global SSL.}
\label{fig:cka_simple}
\end{figure}
\textbf{Representational similarity analysis.} Using Centered Kernel Alignment (CKA)  \citep{kornblith2019similarity}, we study the representations learned by the signal encoder on Dataset 2 under different SSL methods. Our findings are: (1) representations from \ourmethodshort encode aspects of both component-only SSL and global-only SSL (illustrated in Figure \ref{fig:cka_simple}); (2) the component loss appears to have more effect in the earlier layers of the signal encoder, whereas the global loss has more effect in later layers (details in Appendix \ref{app:sec:results}).

\section{Scope and Limitations} 
\textbf{Intended use-case.} Our method is most appropriate in settings where we have  multimodal trajectory data for patients, i.e., both structured data and ECGs are available. This use case is driven by our target application of monitoring patients with cardiovascular disease (Dataset 1), where the majority of patients have structured data and ECGs. 
In datasets where a small proportion of patients have multimodal data (such as in Dataset 2), it may be preferable to use other unimodal SSL approaches so that patients without multimodal data can still be included in model development. Although Dataset 2 does not exactly match our intended use case, we still considered it because it is publicly available and well-studied in related work.

\textbf{Choice of data augmentations.} \ourmethodshort uses data augmentations to generate different views. Our main contribution is in our two-level loss function, so we did not investigate novel augmentation strategies, instead leveraging existing effective data augmentations for clinical data. Our framework could equally apply with different augmentations or multiview generation approaches.

\textbf{Additional data modalities.} Our experiments focused on clinical time series consisting of a sequence of structured data and high-dimensional physiological signals; we did not include predictive information that may be available from other modalities, such as medical imaging. \ourmethodshort could be readily extended to this scenario, and it could be a valuable direction to explore. 

\section{Conclusion}
\label{sec:conclusion}
In this work, we outlined a self-supervised learning (SSL) strategy for complex clinical time series where individual timesteps in the sequence also contain high-dimensional information, such as physiological signals. 
Our method, Sequential Multi-Dimensional SSL (\ourmethodshort), encourages effective pre-training on both a component level (level of individual signals) and a global level (level of the entire sequence). In experiments on two clinical datasets, pre-training with \ourmethodshort and then fine-tuning improves performance on downstream tasks compared to baselines.  Future work could extend \ourmethodshort's component level loss into multiple levels by adopting frameworks from recent work \citep{pmlr-v151-tonekaboni22a, bardes2022vicregl}. This could induce more structure in the pre-training phase, potentially improving performance.

\textbf{Social impact.} Our contribution in this work is mostly methodological. However, given that our application domain is in medicine, a high-risk setting, our method must be thoroughly validated in larger retrospective and prospective studies before any real-world use. This is to understand any potential risks from its use in practice.

\section*{Acknowledgements} 
This work was supported in part by funds from Quanta Computer, Inc. The authors thank the members of the Clinical and Applied Machine Learning group at MIT, the members of the Computational Cardiovascular Research Group at MIT, Paige Stockwell, Neel Dey, and the reviewers for helpful feedback and assistance with the work.

\bibliography{example_paper}
\bibliographystyle{icml2023}

\clearpage

\clearpage
\appendix
\onecolumn
\section{Augmentation Functions for \ourmethod}
\label{app:sec:aug}
In this section, we specify more details about augmentation functions used in our method.

\subsection{Augmentation Details}
We form an augmented trajectory by separately augmenting each of the data modalities within the trajectory, using the following approach for each data type.

\textbf{High-frequency signal $s$:} For each signal in the trajectory of length $T$, we form a pair of augmented views by first splitting the signal into two disjoint segments and then applying random masking and noise addition as augmentations to each view independently, similar to the approach used in CLOCS \citep{kiyasseh2021clocs}. The intuition is that two segments of a signal that are close in time should encode similar physiology, and can therefore be considered paired views. Random masking and noise addition are commonly used as time-series augmentations \citep{gopal20213kg, zhang2022self,raghu2022data, iwana2021empirical}. In more detail:
\begin{itemize}[nosep]
    \item Signal splitting: We split a raw signal of 30 seconds into two disjoint 10 second segments for the first phase of the augmentation process.
    \item Random signal masking: we choose a random 25\% of the signal to set to zero -- this was found to overall be more effective than masking proportions of 10\% and 50\%. 
    \item Noise addition: we add Gaussian noise with standard deviation 0.25 to the signal.
\end{itemize}

\textbf{Structured-time series data $w$:} The tabular data sequence forms a $T \times M$ matrix over all timesteps of the trajectory. Following prior work \citep{yeche2021neighborhood}, we apply two data augmentation strategies to this matrix:  history cutout and noise addition. In more detail:
\begin{itemize}[nosep]
    \item History cutout: For each feature, with probability 0.25, randomly set 25\% of the timesteps in that timeseries to be missing. Forward fill impute this value. This mirrors the imputation strategy used in our raw data. Unlike in \citet{yeche2021neighborhood}, we use forward filling rather than replacing with zeros, because our time series are much shorter (8 timesteps rather than 48) and therefore replacing with zeros destroyed too much information. Using more aggressive cutout augmentations was found to worsen performance, likely because they destroyed too much information in the data. This is in general a challenge when using relatively short time series.
    \item Noise addition: For each feature, add Gaussian noise with standard deviation equal to 10\% of the standard deviation of that feature's values in the training set.
\end{itemize}

\textbf{Static features $d$:} Following \citet{yeche2021neighborhood}, we use random dropout and noise addition. Other corruption strategies (e.g., \citet{bahri2021scarf}), were found to be less effective, potentially due being too strong  (also seen in \citet{levin2022transfer}). We randomly drop out 25\% of the features (impute with the mean value) and add add Gaussian noise with standard deviation equal to 10\% of the standard deviation of that feature's values in the training set.

\subsection{Other Augmentations} 
Early on in our experiments, we investigated other augmentation functions such as channel dropout for the structured time-series data \citep{yeche2021neighborhood} and more complex signal augmentations, such as random lead masking \citep{oh2022lead}. However, we found the improvements from these to be inconsistent and the hyperparameters to be difficult to tune, so we opted for this more focused set of augmentations.

\section{Further Experimental Details}
In this section, we provide further details about our experiments. We first describe more about the datasets and data preprocessing. We then provide further information on the model architecture for our method and baselines, and discuss our hyperparameter search and settings. We then provide additional quantitative results and representational similarity analysis. 

\subsection{Dataset Details}
\label{app:sec:dataset_details}
As discussed in the main text, we consider two clinical datasets in our experiments:
\begin{itemize}[nosep]
    \item \textbf{Dataset 1} is a private dataset derived from the electronic health record (EHR) of the Massachusetts General Hospital (MGH), consisting of a cohort of patients with a prior diagnosis of heart failure.    
    For each patient, we have structured data from the EHR and physiological signals measured by a bedside telemetry monitor. These signals include vitals signs such as heart rate (HR) and oxygen saturation (SpO2), measured at a low frequency (0.5 Hz), and waveforms such as the electrocardiogram (ECG), measured at a high frequency (240 Hz).

    This dataset was obtained with IRB approval (protocol number 2020P003053). Since the dataset has some identifiable information, all computations are performed on a server that sits behind the hospital firewall. Due to restrictions surrounding its use, this dataset cannot be released at this stage.
    
    \item \textbf{Dataset 2} is a public dataset derived from the commonly used MIMIC-III clinical database \citep{johnson2016mimic, Goldberger2000PhysioBankPA} and its associated database of physiological signals \citep{Moody2020-iq}. The clinical database contains structured data over a patient's stay, and the physiological signals database contains vitals signs (HR, SpO2) and waveforms (ECG) measured by a bedside telemetry monitor. 
    
    We use the widely adopted preprocessing pipeline introduced in \citet{harutyunyan2019multitask} to form the specific cohort and extract the structured data features used in modeling. This pipeline also provides the functionality to create development and testing sets for the different downstream tasks we consider. 

    The clinical database is available on PhysioNet \citep{Goldberger2000PhysioBankPA} to credentialed users. The database of physiological signals is open-access on PhysioNet.
\end{itemize}

\begin{table*}[t]
\centering
\caption{Dataset statistics.}
\begin{tabular}{@{}lcccc@{}}
\toprule
                                   & \multicolumn{2}{c}{Dataset 1}                      & \multicolumn{2}{c}{Dataset 2} \\ \midrule
\multicolumn{1}{l|}{Task}          & \# Patients & \multicolumn{1}{c|}{\# Trajectories} & \# Patients & \# Trajectories \\ \midrule
\multicolumn{1}{l|}{Pre-Training}  & 8888        & \multicolumn{1}{c|}{43858}           & 5022        & 26615           \\
\multicolumn{1}{l|}{Elevated mPAP} & 2025        & \multicolumn{1}{c|}{48511}           & 500         & 14957           \\
\multicolumn{1}{l|}{24hr mortality}                    & 9605        & \multicolumn{1}{c|}{57758}                                & 5689        & 318306          \\ \bottomrule
\end{tabular}
\label{app:tab:dataset-stats}
\end{table*}

\textbf{Constructing PT and FT sets.} As outlined in Section \ref{sec:datasets-tasks}, both datasets consist of a number of hospital visits, which we resample at hourly resolution. We extract 30 seconds of the high-dimensional physiological signals at each hour marker from the raw data store.

To generate PT trajectories from these resampled visits, we first split each visit into non-overlapping contiguous 12 hour blocks. A PT trajectory is formed by first sampling a 12 hour block from all the extracted blocks, and then selecting 8 contiguous timesteps from the sampled block (with the starting timestep selected randomly). This trajectory construction strategy has implications in terms of the negative samples in both losses:
\begin{itemize} [nosep]
    \item \textbf{Global loss.} Consider a sampled anchor trajectory from a given patient $i$, timesteps $1-8$. When computing the global loss, the negative pairs for that anchor trajectory are either: (1) a trajectory from a different patient $j \ne i$; or (2) or a trajectory from that same patient $i$ starting after timestep 8. 
    \item \textbf{Component loss.} When computing the component loss, recall that for an anchor signal, other signals from the same trajectory are \textit{not} used as negatives, in order to minimize correlation between the anchor and negatives. Therefore, negative pairs for an anchor signal are either signals from a different patient, or signals from that same patient from further off in time. 
\end{itemize}
This strategy of sampling trajectories that do not overlap was applied to ensure that we do not use highly correlated trajectories/signals as negatives. We note that this strategy is not necessarily optimal, and that different approaches could be used for both the component and global loss terms. Our framework could easily be used with these other sampling strategies and loss formulations.

To generate FT sets, we first use a sliding window to select contiguous 8 hour blocks at 1 hour increments from each visit. Each of these contiguous 8 hour blocks becomes a trajectory in the FT dataset. The trajectory labels are formed based on the nature of the specific task. For example, for the 24 hour mortality task, the label is based on whether the patient dies within 24 hours of the ending time of that trajectory.

\textbf{Trajectory Features and Preprocessing.} 
The trajectories in PT and FT sets consist of static features $d$ and a time-series of structured data and physiological signals $\{(w_t,s_t)\}_{t=1}^{T}$, as presented in Section \ref{sec:methods:probsetup}. 

In Dataset 1, $d \in \mathbb{R}^9$ contains the following features from the EHR: BUN, Chloride, CO2, Creatinine, Glucose, Potassium, Sodium, Systolic Blood Pressure, Diastolic Blood Pressure. We take the average if multiple values are recorded in each time window. $w_t \in \mathbb{R}^{30}$ has mean, standard deviation, maximum, and minimum of heart rate and SpO2 recorded within each hour window (this is sourced from the telemetry monitor), and also 22 heart rate variability features from the ECG recorded by the telemetry monitor during each time window. $s_t \in \mathbb{R}^{4\times2400}$ is a 10 second 4-channel ECG measured at 240 Hz, containing leads I, II, III, and V1, extracted from a longer ECG measured by the telemetry monitor during that hour window.

In Dataset 2, $d\in \mathbb{R}^{38}$ contains the following features from the EHR: FiO2, Glucose, Temperature, pH, and one-hot encoded Glasgow Coma Scale measures, following \citet{harutyunyan2019multitask}. $w_t \in \mathbb{R}^{13}$ contains the following information from the physiological signals database: mean, standard deviation, maximum, and minimum of heart rate and SpO2 recorded within each hour window from the telemetry monitor, diastolic blood pressure, systolic blood pressure, mean blood pressure, heart rate, and SpO2 from the EHR.  $s_t \in \mathbb{R}^{1\times1250}$ is a 10 second 1-channel ECG measured at 125 Hz, containing lead II, extracted from a longer ECG measured by the telemetry monitor during that hour window.

Missing structured data are forward-fill imputed where possible (for example, if part of a time series) and otherwise imputed with the mean over the training dataset. Missing signals are represented with zeros.  We drop any trajectories that have more than 1 timestep with a missing signal. For Dataset 2, we note that by dropping any trajectory with more than 1 timestep with a missing signal, we have a smaller dataset than in \citet{harutyunyan2019multitask}. 

\textbf{Forming labels.} The FT labels for the Elevated mPAP task are formed based on the PA pressure waveform recorded for patients (when available) -- if the mean pressure is over 20 mmHg over a 1 minute period at the final timestep of the trajectory, we assign a binary label of 1, and else 0. For the 24 hour mortality task, we use the recorded time of death recorded in the EHR and if it is less than 24 hours from the end time of the trajectory, we assign a label of 1, and otherwise 0. This is as was done in \citet{harutyunyan2019multitask}.

\subsection{Experimental Setup}
\label{app:sec:exptsetup}
\begin{figure}[h]
\centering
\includegraphics[width=0.6\linewidth]{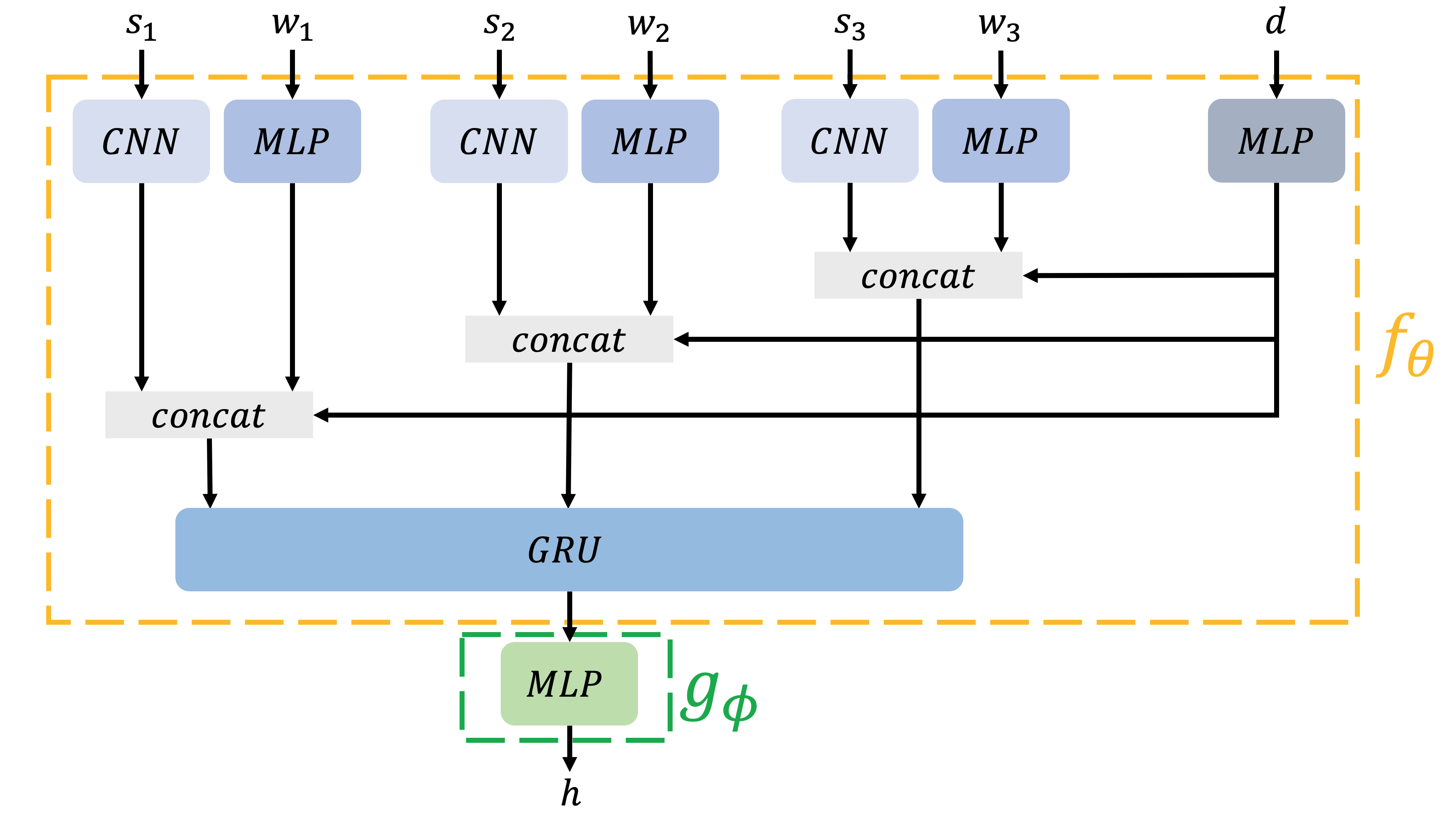}
\caption { \small \textbf{Model architecture used in our experiments.} We show the architecture used to model trajectories in a scenario where the input trajectory has 3 timesteps.}
\label{fig:arch}
\end{figure}

\subsubsection{Model Architecture}
We use the following architecture for the encoder and projection head for all methods:
\begin{itemize}[nosep,]
    \item \textbf{Encoder}: Each signal $s_t$ in the trajectory is passed through a ResNet-styled 1-D CNN encoder with global average pooling. We base our CNN encoder model off a ResNet-18 architecture with kernel size of 15. Following global average pooling over the temporal dimension, each signal is projected into 128 dimensions with a linear layer.
    
    The structured data $w_t$ at each timestep is embedded with a 2-layer fully-connected network with 128 hidden units and ReLU activation at each layer, and this embedding is then concatenated with the signal embedding. The static features $d$ are passed through a different 2-layer fully-connected network with 128 hidden units and ReLU activation at each layer, and then concatenated with the embeddings of the signal and structured data timeseries at each timestep. 
    The resulting sequence of vectors is passed into a 4-hidden layer GRU with hidden size of 384, with the last hidden state of the GRU being used as the overall trajectory embedding vector.
    
    \item \textbf{Projection heads}: The two projection heads for the signal and the trajectory are both  2-layer fully connected networks with batch normalization and ReLU activation with 2048 hidden units. The trajectory projection head takes the last hidden state of the GRU as input, and the signal projection head takes the output of the signals encoder as the input. When using the NT-Xent loss, the resulting projection is normalized \citep{chen2020simple} before computing the NT-Xent loss over the batch. 
\end{itemize}

Figure \ref{fig:arch} shows the model architecture in a scenario in which the input trajectory has 3 timesteps. 

\subsubsection{SSL Methods: Implementation}
We describe the implementation of the SimCLR and VICReg methods in the main paper, Section \ref{sec:methods}. For SimSiam, we follow the setup in \citet{chen2021exploring} and use a predictor network in the trainable branch, and minimize cosine distance between the output of the predictor network in the trainable branch and the output of the projection head in the stop-gradient branch.

For simplicity, we let this predictor network have the same architecture as the projection head -- a two layer fully connected network with batch normalization and ReLU activation, with 2048 hidden units. We did not find a bottleneck structure to improve performance in initial investigations, but further experiments may be warranted here.

\subsubsection{Loss, Architecture, and Optimization Hyperparameters}
There are various hyperparameters to tune, such as learning rates, loss weighting for VICReg loss terms, and loss weighting for \ourmethodshort.
Evaluating many hyperparameter settings is very computationally expensive (since it entails doing both a PT and FT run), so we conduct a reduced search on a subset of the hyperparameters focusing only on the Elevated mPAP task in the unimodal setting, optimizing validation AUROC.

In our hyperparameter search, we use the following setup:
\begin{itemize}[nosep]
    \item Learning rate: tune on a randomly initialized model for the elevated mPAP task on each dataset, and then use this learning rate for all other experiments. We compared Adam with a learning rate of 1e-4, 3e-4, 1e-3, and 3e-3. We found 1e-3 to be the most stable and best performing.
    \item VICReg loss weights: Tune these for the VICReg (global) model only, and use the best hyperparameters for all other uses of the VICReg loss, including \ourmethodshort (VICReg). Following the original paper, we set the covariance weight $\nu =1$ and then tune the invariance weight $\lambda$ and variance weight $\mu$. We found in early experiments that the variance weight did not have much impact on performance, and so focused on the invariance weight, studying $\lambda = 1, 2, 5$. We found $\lambda =1$ to perform the best on both datasets.
    \item \ourmethodshort (SimCLR) component loss weight: Set the global weight $\alpha= 1$ in Eqn. \ref{eqn:multilevel}, and tune the component weight $\beta$, on both datasets separately, comparing $\beta = 0.25, 0.5, 1.0, 2.0$.  We  found $\beta=1.0$ to perform the best on Dataset 1, and $\beta=0.25$ to perform the best on Dataset 2. 
    \item \ourmethodshort (VICReg) component loss weight: Use the best VICReg loss weights found above, set the global weight  $\alpha= 1$ in Eqn. \ref{eqn:multilevel}, and tune the component loss weight $\beta$, comparing $\beta = 0.1, 0.25, 0.5, 1.0, 2.0$.  We  found $\beta=1.0$ to perform the best on Dataset 1, and $\beta=0.1$ to perform the best on Dataset 2. 
\end{itemize}
We fixed the temperature of the NT-Xent loss to 0.1, following \citet{yeche2021neighborhood}.

We did not conduct tuning of the architecture hyperparameters, and instead opted to use architectural choices that were found to be effective in previous works, such as a ResNet signal encoder \citep{raghu2021meta,raghu2022data}, a wide projection head \citep{chen2020big, bardes2022vicreg}, and a GRU sequence model \citep{mcdermott2021chil}. Similar to \citet{mcdermott2021chil}, we did not find a transformer model to be beneficial as the sequence model, though perhaps architectural tuning could improve its performance.

\subsubsection{Compute Details} 
All models were trained on either a single NVIDIA Quadro RTX 8000 or a single NVIDIA RTX A6000 GPU. Pre-training takes about 8 hours on Dataset 1 and about 2 hours on Dataset 2. Fine-tuning on Dataset 1 tasks takes about 4 hours. Fine-tuning in Dataset 2 on Elevated mPAP takes about 30 minutes, and about 4 hours on 24hr Mortality. Pre-training uses approximately 20 GB of GPU memory, and fine-tuning uses approximately 10 GB of GPU memory.

\subsection{Additional Results}
\label{app:sec:results}

\begin{figure}[t]
\centering
\includegraphics[width=0.8\linewidth]{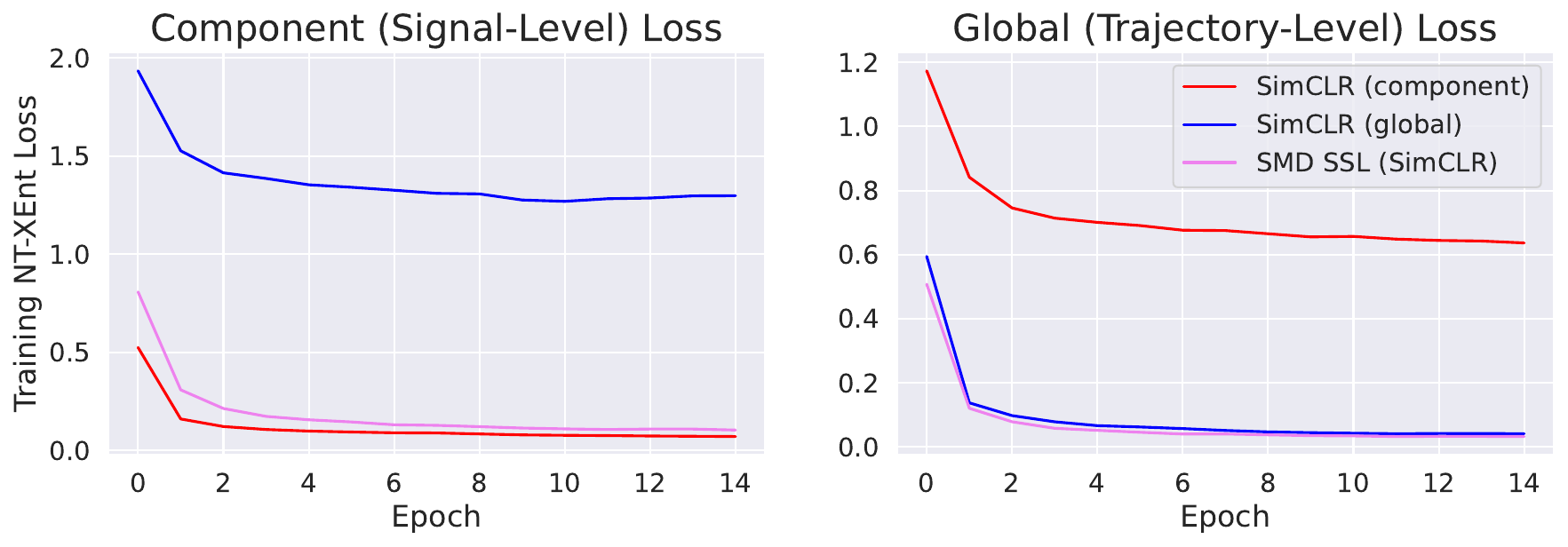}
\caption { \small \textbf{Studying training loss curves for \ourmethodshort and variations.} We observe that \ourmethodshort effectively minimizes both the component and global NT-XEnt losses during training. Interestingly, we observe that the NT-Xent loss computed on the signal level and trajectory level reduces somewhat even when it is not explicitly minimized. To see this, consider the SimCLR (global) method and the component Loss -- the component loss reduces over training even with a random projection head, without adding this term to the objective. This indicates that the global loss and component loss are not entirely independent (as is expected).}
\label{fig:train_curve}
\end{figure}

\textbf{Studying loss curves.} Figure \ref{fig:train_curve} shows training loss curves for SimCLR-based models on Dataset 2. We observe that \ourmethodshort effectively minimizes both component and global losses over training. Considering the component loss alone (left plot), we see that this loss naturally reduces during training of the SimCLR (global) model even though this is not explicitly enforced during model training -- we compute the component loss in this case with a randomly initialized signal level projection head that is not updated, and the network parameters are also not updated to minimize this component loss. An analogous situation with SimCLR (component) and the global loss is seen in the right plot. This suggests that training with one of the losses does encourage structure in both representation spaces, even with random projections, but this structure is more clearly defined when the loss is explicitly minimized (as in \ourmethodshort).

\textbf{Trends in AUPRC.} Since the mortality task has low prevalence, we study trends in AUPRC among methods as they compare to AUROC. We find that in the unimodal setting, on both datasets, our objective improves AUPRC by 1-2\% over baselines, but AUPRCs are all relatively low ($<$5\%) due to the limited predictive signal in the ECGs alone for the mortality prediction task. 
In the multimodal setting, on Dataset 1, SMD SSL worsens AUPRC over the single-level approach by 2.9\% AUPRC (10.8\% vs 7.9\%) -- this is consistent with what was seen with AUROC. On Dataset 2, the AUPRC with our approach is 28.4\%, an improvement of about 4\% over the best baseline.

\begin{table*}[t]
\centering
     \caption{\small Mean and 95\% confidence interval of AUROC on fine-tuning tasks in the unimodal setting when only using structured data, comparing no PT (RandInit) to PT with the SimCLR and VICReg global losses. We find that when considering structured data alone, PT does not offer much benefit to performance; however, there are improvements seen in the Elevated mPAP task.}
     \label{tab:results-unimodal-tab}
    \begin{subtable}[h]{0.5\textwidth}
    \centering
    \caption{\small Results on Dataset 1.}
        \label{tab:dataset1-unimodal-tab}
        \begin{tabular}{@{}lcc@{}}
            \toprule
                         & Elevated mPAP & 24hr Mortality\\ \midrule
            RandInit     & 65.3 $\pm$ 0.1   &\textbf{79.0 $\pm$ 0.4}     \\
            SimCLR       & \textbf{66.8 $\pm$ 0.1}   & \textbf{79.0 $\pm$ 0.4}   \\
            VICReg       & 66.0 $\pm$ 0.0   & 77.9 $\pm$ 0.4     \\ \bottomrule
        \end{tabular}
    \end{subtable}%
    \begin{subtable}[h]{0.5\textwidth}
    \centering
    \caption{\small Results on Dataset 2.}
        \label{tab:dataset2-unimodal-tab}
        \begin{tabular}{@{}lcc@{}}
        \toprule
                     & Elevated mPAP & 24hr Mortality \\ \midrule
        RandInit     &   65.0 $\pm$ 0.3 &  \textbf{90.1 $\pm$ 0.1 }              \\
        SimCLR       &   66.8 $\pm$ 0.3 &  88.1 $\pm$ 0.1              \\
        VICReg       &   \textbf{68.1 $\pm$ 0.3} &  89.3 $\pm$ 0.1  \\ \bottomrule
        \end{tabular}
     \end{subtable}
\end{table*}

\textbf{Unimodal experiments with structured data.} Table \ref{tab:results-unimodal-tab} shows results when training on only the structured data (structured time-series and static vector) in the trajectory. We find that the Elevated mPAP task can benefit from pre-training, but the 24hr Mortality task performance is not boosted by pre-training. This is likely because the structured data are relatively simple and low-dimensional, and there is enough data to learn useful predictive information from these data as-is, without pre-training. 

\textbf{Additional baselines.} As discussed in the main text, related SSL strategies for time series data are not exactly applicable in our setting since they formulate pipelines for structured data-only time series (rather than multimodal time series), or are concerned with individual physiological waveforms (rather than sequences of waveforms). 

Despite these differences, for completeness, we study here the performance of adapted versions of three related methods from the literature for Dataset 2 (MIMIC-III), focusing on the multimodal setting. Specifically, we evaluate the SSL objective functions proposed in NCL \citep{yeche2021neighborhood}, SACL \citep{cheng2020subject}, and CLOCS (specifically the CMSC formulation) \citep{kiyasseh2021clocs}. We use each of these losses in a global-only SSL setup in order to compare how they perform to our proposed two-level loss function. We additionally evaluate structured data-only NCL.

We evaluate these methods using the same experimental setup (augmentation pipeline, optimization hyperparameters, model architecture, etc) as what was used when evaluating our method. For NCL, we considered two values of $\alpha$ (0.3 and 0.5), and fixed $w=16$ as in the original paper’s configuration on MIMIC. We select the value of $\alpha$ that obtained the best validation AUROC on each downstream task.

Results are shown in Table \ref{app:tab:baselines}. As seen, the best two-level approach, SMD SSL (VICReg), outperforms the different baselines. This indicates that a two-level loss is not easily outperformed by other global-only loss functions. An important investigation is to conduct a more thorough hyperparameter search for these alternative loss functions, and also evaluate whether two-level versions of these other objectives could improve on SMD SSL (VICReg).

\begin{table}[t]
\centering
\caption{\small Test AUROC of different SSL algorithms on Dataset 2 (MIMIC) in the multimodal setting. We observe that SMD SSL (VICReg) improves on these additional SSL baselines.}
\begin{tabular}{@{}lcc@{}}
\toprule
                           & Elevated mPAP & 24hr Mortality \\ \midrule
SMD SSL (VICReg)           & 71.6          & 90.7           \\
NCL (Structured data only) & 67.6          & 87.7           \\
NCL (Multimodal)           & 65.5          & 89.9           \\
SACL                       & 64.7          & 89.8           \\
CLOCS                      & 64.3          & 89.9           \\ \bottomrule
\end{tabular}
\label{app:tab:baselines}
\end{table}

\textbf{Linear Evaluation vs Full Fine-tuning.} As discussed in the main paper, our goal is to develop a self-supervised pre-training algorithm that finds an effective model initialization for adaptation to downstream tasks (i.e., a transfer learning setting). As a result, we evaluate both full FT and linear evaluation, reporting the evaluation strategy that obtains the best validation AUROC on a per-method and per-task basis. It is important to consider full FT since it almost always outperforms linear evaluation -- we observed this in our results, and a similar finding was see in the evaluation from \citet{mcdermott2021chil}. Table \ref{app:tab:eval_strat} highlights this finding for a subset of the methods on Dataset 2 (MIMIC), in the multimodal setting. 

\begin{table}[t]
\centering
\caption{\small Comparing test AUROC of the two evaluation paradigms --- Linear Evaluation and Full Fine-tuning (FT) --- 
with selected SSL algorithms on Dataset 2 (MIMIC) in the multimodal setting. We find that Full FT routinely performs better than linear evaluation.}
\begin{tabular}{@{}lcc|cc@{}}
\toprule
                                         & \multicolumn{2}{c|}{Elevated mPAP} & \multicolumn{2}{c}{24hr Mortality} \\ \midrule
\multicolumn{1}{l|}{}                    & Full FT     & Linear Evaluation    & Full FT     & Linear Evaluation    \\ \midrule
\multicolumn{1}{l|}{RandInit}            & 65.3        & N/A                  & 87.8        & N/A                  \\
\multicolumn{1}{l|}{SMD SSL (VICReg)}    & 71.6        & 65.0                 & 90.7        & 72.2                 \\
\multicolumn{1}{l|}{VICReg (Global)}     & 70.4        & 64.4                 & 87.8        & 71.7                 \\
\multicolumn{1}{l|}{SimCLR (Global)}     & 63.7        & 63.1                 & 86.8        & 71.7                 \\
\multicolumn{1}{l|}{SimSiam (Component)} & 67.4        & 61.6                 & 90.6        & 71.9                 \\
\multicolumn{1}{l|}{SimSiam (Global)}    & 60.6        & 50.6                 & 90.4        & 50.0                 \\ \bottomrule
\end{tabular}
\label{app:tab:eval_strat}
\end{table}

\textbf{Studying longer Pre-training.} On Dataset 2 (MIMIC), we pre-trained methods for longer (50 epochs) and compared performance after 15 and 50 epochs, following the best of full FT and linear evaluation (following our standard experimental setup). Results are in Table \ref{app:tab:longpt}, indicating that performance did not improve following longer PT. 

\begin{table}[t]
\centering
\caption{\small Comparing test AUROC after different amounts of PT 
with selected SSL algorithms on Dataset 2 (MIMIC) in the multimodal setting. Performance does not appear to improve after more PT.}
\begin{tabular}{@{}lcccc@{}}
\toprule
                                         & \multicolumn{2}{c}{Elevated mPAP}                & \multicolumn{2}{c}{24hr Mortality} \\ \midrule
\multicolumn{1}{l|}{}                    & 15 epochs PT & \multicolumn{1}{c|}{50 epochs PT} & 15 epochs PT     & 50 epochs PT    \\ \midrule
\multicolumn{1}{l|}{SMD SSL (VICReg)}    & 71.6         & \multicolumn{1}{c|}{66.6}         & 90.7             & 89.3            \\
\multicolumn{1}{l|}{VICReg (Global)}     & 70.4         & \multicolumn{1}{c|}{63.2}         & 87.8             & 78.8            \\
\multicolumn{1}{l|}{SimSiam (Component)} & 67.4         & \multicolumn{1}{c|}{59.6}         & 90.6             & 78.8            \\
\multicolumn{1}{l|}{SimSiam (Global)}    & 60.6         & \multicolumn{1}{c|}{51.1}         & 90.4             & 88.2            \\ \bottomrule
\end{tabular}
\label{app:tab:longpt}
\end{table}

\textbf{Additional representational similarity experiments.} In the main text, we presented a simple representational similarity study examining the how the learned representations by the CNN signals encoder compared in \ourmethodshort (SimCLR) vs. SimCLR (component only) and SimCLR (global only) pre-training. We found that \ourmethodshort representations had reasonable Centered Kernel Alignment (CKA) similarity \citep{kornblith2019similarity} with both component-only and global-only PT. On the other hand, global-only and component-only PT were quite dissimilar. 

\begin{figure}[t]
\centering
\includegraphics[width=0.5\linewidth]{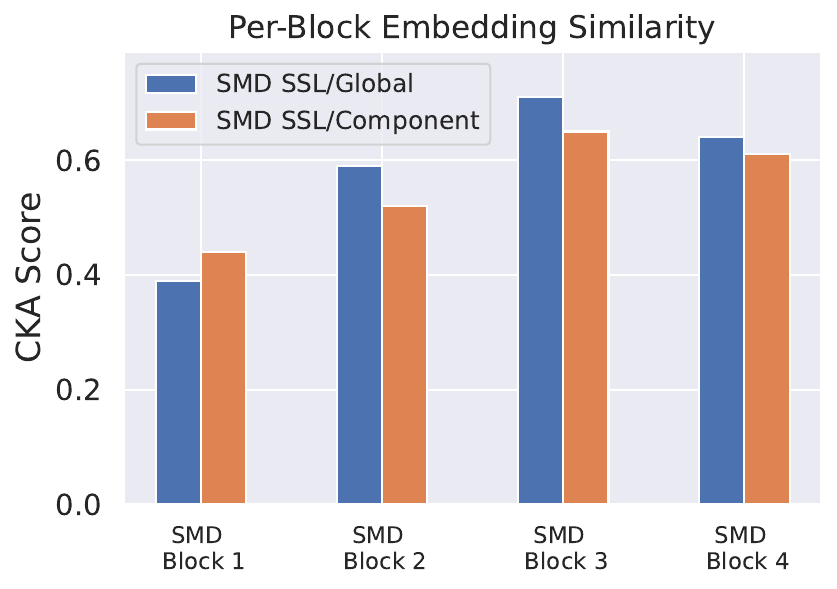}
\caption { \small \textbf{Studying per-block representational similarity in the CNN encoder between \ourmethodshort pre-training and component-only and global-only pre-training.} \ourmethodshort representations are more similar to component-only PT early on in the CNN signals encoder, and more similar to global-only PT deeper in the network.}
\label{fig:cka_complex}
\end{figure}

To further understand the effect of training with the component and global losses in \ourmethodshort, we conduct a finer-grained CKA study. We take the output of each residual block of the CNN signals encoder and compare the CKA similarity in these representations (following pooling over the sequence dimension) to the CKA similarity of component-only and global-only PT models, on a per-block basis. That is, we average the CKA similarity between \ourmethodshort (block $i$) and component-only PT (blocks $1, 2, 3, 4$), and similarly for global-only PT. The results are shown in Figure \ref{fig:cka_complex}. We see that \ourmethodshort PT has more similarity with component-only PT in the first block, and greater similarity with global-PT in the remaining blocks. This suggests that the component loss is having the most impact on \ourmethodshort representations in the earlier CNN layers, indicating that these low-level features are particularly relevant for minimizing the component loss -- this makes sense, since we would expect lower-level features to matter more in a per-signal embedding, and global-level features to matter more in a sequence-level embedding.

\end{document}